\documentclass[epsf,fleqn]{aicom2e}
\usepackage{latexsym} % needed for \Box in \punto
\usepackage{epsfig}
\usepackage{amssymb}
\usepackage{url}
\usepackage{xspace}
\newcommand{\HELP}{\vspace*{-4mm}}
\newcommand{\refFig}[1]{Figure~\ref{fig:#1}}
\newcommand{\cmpitem}{\ensuremath{\bullet}\xspace}
\newcommand{\nop}[1]{}

\newcommand{\dlv}{{\bf\small{DLV}}}
\nop{
\newtheorem{theorem}{Theorem}[section]
\newtheorem{example}[theorem]{Example}

\newtheorem{definition}[theorem]{Definition}

}
\newcommand{\derives}{\mbox{\,\texttt{:\hspace{-0.15em}-}}\,\xspace}
\newcommand{\Or}{\ensuremath{\mathtt{\,v\,}}\xspace}

\newcommand{\pr}[1]{\stepcounter{#1}(\arabic{#1})}
\newcommand{\dlpp}{{\sc DLP}\ensuremath{^{\small \bigvee,\bigwedge}}\xspace}
\newcommand{\p}{\ensuremath{{\cal P}}}

\newcommand{\R}{\ensuremath{r}}

\newcommand{\dlrule}[2]{\ensuremath{#1 \derives #2.}}
\newcommand{\naf}{\ensuremath{\mathrm{not}}\xspace}

\newcommand{\aprog}[1][]{\ensuremath{\p_{#1}}}

\newcommand{\tuple}[1]{\langle#1\rangle}
\newcommand{\NP}{{\rm NP}}
\newcommand{\GP}{\ensuremath{Ground(\p)}\xspace}
\newcommand{\BP}{\ensuremath{B_{\p}}\xspace}
\newcommand{\UP}{\ensuremath{U_{\p}}\xspace}

\newcommand{\SigmaP}[1]{{\Sigma}_{#1}^{P}}

\newcommand{\LS}{\ensuremath\{$L: Conj$\}}
\newenvironment{dlvcode}
  {\begin{displaymath}\begin{array}{l}}
  {\end{array}\end{displaymath}}
\newcommand{\DLP}{{\sc DLP}\xspace}

\newcommand{\gc}{{\bf \small{GC}}\xspace}

\newcommand{\factprog}{\ensuremath{{\cal F}}}
\newcommand{\guessprog}{\ensuremath{{\cal G}}}
\newcommand{\checkprog}{\ensuremath{{\cal C}}}

\newcommand{\code}[1]{\ensuremath{#1}}

\newcommand{\HR}{\ensuremath{H(\R)}}
\newcommand{\BR}{\ensuremath{B(\R)}}
\newcommand{\dlconstraint}[1]{\ensuremath{\derives #1.}}

\def\gp{{\cal P}}
\newcommand{\Comma}{\,,\ }

\newcommand{\etal}{\textit{et al.}\xspace}

\newtheorem{mdefinition}{Definition}
\newtheorem{mexample}{Example}

\newenvironment{example}{\begin{mexample}}{\hfill$\triangle$\end{mexample}}
\newenvironment{definition}{\begin{mdefinition}}{\hfill$\Box$\end{mdefinition}}
\begin{document}
\begin{frontmatter}                           % The preamble begins here.
%
%\pretitle{Pretitle}
\title{Parametric Connectives in\\
Disjunctive Logic Programming} \runningtitle{Parametric
Connectives in Disjunctive Logic Programming}
%\subtitle{Subtitle}
\maketitle
\author{\fnms{Simona} \snm{Perri}}
and
\author{\fnms{Nicola} \snm{Leone}}
\runningauthor{S. Perri et al.}
\address{Department of Mathematics,\\
University of Calabria\\
I-87030 Rende (CS), Italy\\
E-mail: \{leone,perri\}@mat.unical.it}

\begin{abstract}
Disjunctive Logic Programming (\DLP) is an advanced formalism
for Knowledge Representation and Reasoning (KRR).
\DLP is very expressive in a precise mathematical sense:
it allows to express every property of finite structures
that is decidable in the complexity class $\SigmaP{2}$ ($\NP^{\NP}$).
Importantly, the \DLP encodings are often simple and natural.

In this paper, we single out some limitations of \DLP for KRR,
which cannot naturally express problems where the size of the
disjunction is not known ``a priori'' (like N-Coloring), but it is
part of the input. To overcome these limitations, we further
enhance the knowledge modelling abilities of \DLP, by extending
this language by {\em Parametric Connectives (OR and AND)}. These
connectives allow us to represent compactly the
disjunction/conjunction of a set of atoms having a given property.
We formally define the semantics of the new language, named \dlpp
and we show the usefulness of the new constructs on relevant
knowledge-based problems.
We address implementation issues and discuss related works.
\end{abstract}

\end{frontmatter}

\section{Introduction}\label{sect:intro}
Disjunctive logic programs are logic programs where disjunction is
allowed in the heads of the rules and negation may occur in the bodies
of the rules.
Such programs are now widely recognized as a valuable
tool for knowledge representation and commonsense reasoning
\cite{bara-gelf-94,lobo-etal-92,IFIP-94,eite-etal-99-lbai,gelf-lifs-91,lifs-96,mink-94,bara-2002}.
The most widely accepted semantics for \DLP is the
{\em answer sets semantics} proposed by Gelfond and Lifschitz
\cite{gelf-lifs-91} as an extension of the stable model semantics of
normal logic programs \cite{gelf-lifs-88}.  According to this
semantics, a disjunctive logic program may have
several alternative models (but possibly none), called {\em answer
sets}, each corresponding to a possible view of the world.
Disjunctive logic programs under answer sets semantics are very
expressive.
It was shown in \cite{eite-etal-97f,gott-94b} that, under
this semantics, disjunctive logic programs capture the complexity
class $\SigmaP{2}$ (i.e., they allow us to express, in a precise
mathematical sense, {\em every} property of finite structures over a
function-free first-order structure that is decidable in
nondeterministic polynomial time with an oracle in $\NP$).  As Eiter
\etal \cite{eite-etal-97f} showed, the expressiveness of disjunctive
logic programming has practical implications, since relevant practical
problems can be represented by disjunctive logic programs, while they
cannot be expressed by logic programs without disjunctions, given
current complexity beliefs.
Importantly, even problems of lower complexity
can be often expressed more naturally by disjunctive programs
than by programs without disjunction.

As an example, consider the well-known problem of 3-coloring,
which is the assignment of
three colors to the nodes of a graph in such a way that adjacent
nodes have different colors. This problem is known to be
NP-complete. Suppose that the nodes and the edges are
represented by a set $F$ of facts with predicates $node$ (unary)
and $edge$ (binary), respectively. Then, the following \DLP
program allows us to determine the admissible ways of coloring
the given graph.
{\small
\begin{dlvcode}
%r_1:\quad color(X,red) \Or color(X,yellow) \Or color(X,green)
\hspace{-0.7cm}r_1:\ \ color(X,r) \Or color(X,y) \Or color(X,g)
\derives node(X).\\
\hspace{-0.7cm}r_2:\ \ \derives edge(X,Y), color(X,C), color(Y,C).
\end{dlvcode}
}
Rule $r_1$ above states that every node of the graph is
colored {\bf r}ed or {\bf y}ellow or {\bf g}reen,
while $r_2$ forbids the assignment of
the same color to any adjacent nodes.
The minimality of answer sets guarantees that every node is assigned
only one color.
Thus, there is a one-to-one correspondence between the solutions
of the 3-coloring problem and the answer sets of $F\cup\{r_1, r_2\}$.
The graph is 3-colorable if and only if $F\cup\{r_1, r_2\}$ has
some answer set.

\vspace{0.2cm}
Despite the high expressiveness of \DLP,
there are several problems which cannot be encoded in \DLP in a simple
and natural manner.
Consider, for instance, the generalization of the 3-coloring
problem above, where the number of admissible colors is not
known ``a priori'' but it is part of the input.
This problem is called {\em N-Coloring}:
Given a graph {\em G} and a set of {\em N} colors,
find an assignment of the {\em N} colors to the nodes of {\em G}
in such a way that adjacent nodes have different colors.

The most natural encoding for this problem
would be obtained by modifying rule $r_1$ in the above encoding
of 3-coloring.
The head
{\small
\begin{dlvcode}
color(X,r) \Or color(X,y) \Or color(X,g)
\end{dlvcode}
}
should be replaced by a disjunction of $N$ atoms
representing the $N$ possible ways of coloring the node at hand.
This encoding, however, cannot be done in a uniform way,
since the number of colors is not known ``a priori''
but it is part of the input (the program should be changed
for each number $N$ of colors; while a uniform encoding
requires the program to be fixed, and only the facts
encoding the input to be varying).

To overcome these limitations, in this paper
we enhance the knowledge modelling abilities of \DLP,
by extending this language by
{\em Parametric Connectives (OR and AND)}.
These connectives allow us to represent compactly the disjunction/conjunction
of a set of atoms having a given property.
For instance, by using parametric OR
we obtain a simple and natural encoding
of N-Coloring by modifying the above rule $r_1$ as follows:
{\small
\begin{dlvcode}
\bigvee\{col(X,C):  col(C) \} \derives node(X).
\end{dlvcode}
}
(see Section \ref{sec:N-Coloring}).
Intuitively, if the input colors are given by facts $col(c_1),\cdots,col(c_n)$, then the above rule stands for
{\small\begin{dlvcode}col(X,c_1) \Or \cdots \Or col(X,c_n) \derives node(X)\end{dlvcode}}
Shortly, the main contribution of the paper are the following \\
\cmpitem
We extend Disjunctive Logic Programming by parametric connectives
and formally define the semantics of the resulting language, named \dlpp.\\
\cmpitem
We address knowledge representation issues, showing the impact
of the new constructs on relevant KR problems.\\
\cmpitem
We discuss some implementation issues, providing the design of an extension of the \dlv\ system to support \dlpp.

The sequel of the paper is organized as follows.
In Section \ref{sec:language}, we provide the syntax and the semantics of
the \dlpp language.
In Section \ref{sec:gcc}, we illustrate a methodology for declarative programming in standard \DLP.
In Section \ref{sec:kr}, we address knowledge representation issues in \dlpp.
In Section \ref{sec:implementation}, we describe the implementation of the \dlpp language in the
\dlv\ system.
In Section \ref{sec:relwork}, we discuss related works.
Finally, in Section \ref{sec:conclusion}, we draw our conclusions.

\section{The \dlpp Language}
\label{sec:language} In this section, we provide a formal
definition of the syntax and the semantics of the \dlpp language.
\subsection{Syntax}
A variable or a constant is a {\em term}.  A {\em standard} {\em
atom} is $a(t_{1},..., t_{n})$, where $a$ is a {\em
predicate} of arity $n$ and $t_{1},..., t_{n}$ are terms.  A
{\em standard~literal} is either a {\em standard~positive~literal}
$p$ or a {\em standard~negative~literal} $not\ p$, where
$p$ is a standard atom. A {\em standard conjunction} is $k_1,\cdots,k_n$
where each $k_1, \cdots, k_n$ is a standard
literal.
A {\em symbolic literal set} $S$ is \LS, where $L$ is a
standard literal and $Conj$ is a standard conjunction; $L$ is called the parameter of $S$ and $Conj$
is called the domain of $S$;
if $L$ is a positive standard literal, $S$ is called {\em positive symbolic literal set}.
A {\em parametric~AND~literal} is $\bigwedge S$ where $S$ is a symbolic literal set. A {\em parametric~OR~literal}
is $\bigvee S$ where $S$ is a positive symbolic literal set.

\begin{example}\label{ex:paratom}\em
{\small
$\bigvee \{ a(X,Y): q(X,Y), not \ r(Y)\}$
} is a parametric OR literal and {\small$\{ a(X,Y): q(X,Y), not \ r(Y)\}$}
is the positive symbolic literal set.
Intuitively, the above parametric OR literal stands for the disjunction of all instances of $a(X,Y)$ such that
the conjunction $q(X,Y), not \ r(Y)$ is true.
\end{example}

 A {\em (disjunctive) rule} $r$ is a
syntactic of the following form: \vspace{-0.2 cm}
\[
a_1\Or\cdots\Or a_n\derives l_1,\cdots, l_m. \qquad n\geq0,
m\geq0
\]

where $a_1,\cdots ,a_n$ are standard positive literals or parametric~OR~literals and $l_1, \cdots, l_m$ are
standard literals or parametric~AND~literals.\\
The disjunction $a_1\Or\cdots\Or a_n$ is the {\em head} of $r$,
while the conjunction $l_1,\cdots, l_m$ is the {\em body} of $r$.

We denote by $H(r)$ the set \{$a_1 ,..., a_n$\} of the head literals,
and by $B(r)$ the set \{$l_1 ,..., l_m$\} of the body literals.
An \textit{(integrity) constraint} is a rule with an empty head.

A {\em \dlpp program} $\p$ is a finite set of rules. A $\neg$-free
(resp., $\Or$-free) program is called {\em positive}\/ (resp.,
{\em normal}). A program where no
parametric literals appear is called ({\em standard}) {\em DLP
program}. A term, an atom, a literal, a rule, or a program are
{\em ground} if no variables appear.
%A ground program is also called a {\em propositional} program.

\subsection{Syntactic Restrictions and Notation}

A variable $X$ appearing solely in
% a symbolic set $\mathit{Vars} : Conj$ of
a parametric literal of a rule $r$ is a {\em local variable} of $r$.
The remaining variables of $r$ are called {\em global variables} of $r$.
%in particular, local variables in $\mathit{Vars}$ are {\em bound} local variables,
%while the remaining local variables are {\em unbound} local variables.
\begin{example}\em
Consider the following rule
{\small
\begin{dlvcode}
p(Y,Z) \derives \bigwedge \{ q(X,Y) : a(X,Z))\}, t(Y),r(Z).
\end{dlvcode}
}

$X$ is the only local variable, while $Y$ and $Z$ are global variables.
\end{example}
% Safety Restrictions.
%In the following we describe the safety of the aggregate atoms and
%the stratification of \DLPA programs.
\subsubsection*{Safety} A rule $r$ is {\em safe} if the following conditions hold:
\begin{itemize}
\item[(i)] each global variable of $r$ appears in a positive
standard literal occurring in the body of $r$; \item[(ii)] each local
variable of $r$ appearing in a symbolic set \LS, also appears in a
positive literal in $Conj$.
\end{itemize}
A program is {\em safe} if all of its rules are safe.
%%%%% FIXME aggregate with assignment. %%%%%
\begin{example}\label{ex:safety}
\em
Consider the following rules:
\[
{\small
\begin{array}{l}
\code{\dlrule{\bigvee \{  p(X,Y): q(Y)\}}{r(X)}}
\vspace{3pt}\\
\code{\dlrule{ p(X,Z)} {\bigwedge \{ q(X,Y) : a(X))\}, s(X,Z)}}
\vspace{3pt}\\
\code{\dlrule{p(X)}{\bigwedge \{ q(X,Y) : a(X)\}, t(Y)}}
\end{array}
}
\]
The first rule is safe, while the second is not, since the local
variable $Y$ violates condition (ii). The third rule is not
safe either, since the global variable $X$ violates condition (i).
\end{example}

\subsubsection*{Stratification}
A \dlpp program \p\ is {\em p-stratified} if there exists
a function $||\ ||$, called {\em level mapping}, from the set of (standard)
predicates of \p\ to ordinals,
such that for each pair $a$ and $b$ of (standard) predicates
of \p, and for each rule $r \in \p$ the following conditions hold:

(i)for each parametric literal $\gamma$ of $r$, if $a$ appears in the parameter of $\gamma$ and $b$ appears in the domain
of $\gamma$ then $||b|| < ||a||$, and \\
(ii) if $a$ appears in the head of $r$, and $b$ occurs in a standard atom
in the body of $r$, then $||b||\leq ||a||$.

\begin{example}\label{ex:stratification}
\em
Consider the program consisting of a set of facts for predicates $a$ and $b$,
plus the following two rules:\\
\begin{dlvcode}
p(X) \derives q(X), \bigwedge\{q(Y) : a(X,Y),b(X)\}. \\
q(X) \derives p(X), b(X).\\
\end{dlvcode}
The program is p-stratified, as the level mapping \ \  $||a||=||b||=1\quad ||p||=||q||=2$ \ \  satisfies
the required conditions.
If we add the rule $b(X) \derives p(X)$,
then no legal level-mapping exists and the program becomes p-unstratified.
\end{example}
From now on, throughout this paper, we assume that all rules of a $\dlpp$ $\p$ are safe and p-stratified.

\subsection{Semantics}
\label{sec:semantics}

{\bf Program Instantiation.} Given a \dlpp program \p, let \UP
denote the set of constants appearing in \p, and \BP the set of
standard atoms constructible from the (standard) predicates of \p\
with constants in \UP.

A {\em substitution} is a mapping from a set of variables to the
set $\UP$ of the constants appearing in the program \p. A
substitution from the set of global variables of a rule $r$ (to
$\UP$) is a {\em global substitution for r}; a substitution from
the set of local variables of a symbolic set $S$ (to $\UP$) is a
{\em local substitution for $S$}. Given a symbolic set without
global variables $S$ = \LS, the {\em
instantiation of set $S$} is the following ground set of pairs
\\
$S' = \{ \tuple{\gamma(L): \gamma(Conj)} \mid$
$\gamma$ {\rm is a local substitution for }$S\}$%
\footnote{Given a substitution $\sigma$ and a \dlpp object $Obj$
(rule, conjunction, set, etc.), with a little abuse of notation,
we denote by $\sigma(Obj)$
the object obtained by replacing each variable $X$ in $Obj$ by $\sigma(X)$.};
$S'$ is called {\em ground literal set}.

A {\em ground instance} of a rule $r$ is obtained in two steps:
(1) a global substitution $\sigma$ for $r$ is first applied over
$r$; (2) every symbolic set $S$ in $\sigma(r)$ is replaced by its
instantiation $S'$. The instantiation \GP\ of a program \p\
is the set of all possible instances of the rules of \p.
\begin{example}\label{ex:instantiation}
\em Consider the following program
$\p_1$:
{\small
\begin{dlvcode}
p(1) \Or q(2,2).
\vspace{3pt}\\
p(2) \Or q(2,1).\vspace{3pt}\\
s(X) \derives p(X), \bigwedge\{a(Y):q(X,Y)\}.
\end{dlvcode}
}
The instantiation $Ground(\p_1)$ is the following:
{\small
\begin{dlvcode}
p(1) \Or q(2,2).
\vspace{3pt}\\
p(2) \Or q(2,1).
\vspace{3pt}\\
s(1) \derives p(1), \bigwedge\{\tuple{a(1):q(1,1)},\tuple{a(2):q(1,2)}\}.
\vspace{3pt}\\
s(2) \derives p(2), \bigwedge\{\tuple{a(1):q(2,1)},\tuple{a(2):q(2,2)}\}.
\end{dlvcode}
}
\end{example}
%We describe this by transforming our syntactic constructions in a different
%domain, therefore we have to define an association between constants and
%objects in the different domain and an association between predicate symbols
%and functions which associate true or false to a given n-tuple.

{\bf Interpretation and models.} An {\em interpretation} for a
\dlpp program \p\ is a set of standard ground atoms $I\subseteq
\BP$.

A ground positive literal $A$ is {\em true} (resp., {\em false})
w.r.t. $I$ if $A \in I$ (resp., $A \not\in I$).
A ground negative literal $\neg A$ is {\em true} w.r.t. $I$ if $A$ is false
w.r.t. $I$; otherwise $\neg A$ is false w.r.t. $I$.

 Besides assigning truth values to the standard ground
literals, an interpretation provides the meaning also to
(ground)literal sets, and to (the instantiation of) parametric literals. Let $S$ be a
(ground) literal set. The valuation $I(S)$ of $S$ w.r.t.\ $I$
is the set
\[
\{ L \ | \ (L:conj \in S) \wedge (conj \ is \ true \ w.r.t \ I)
\}.
\]

Given a parametric OR literal $\bigvee S$, let $S'$ be the instantiation of $S$. Then $\bigvee S'$
is true w.r.t $I$
if at least one of the standard literals in $I(S')$ is true w.r.t
$I$. Similarly, given a parametric AND literal $\bigwedge S$, let $S'$ be the instantiation of $S$.
$\bigwedge S'$ is true w.r.t $I$
if all the standard literals in $I(S')$ are true w.r.t $I$.
\begin{example}\label{ex:f(Q)EI(f(Q))}
\em
Let $\UP$ be the set \{1,2\} and $I$ the interpretation
$\{p(1),p(2), a(1,2), a(2,1), b(1), b(2)\}$. Consider the parametric
AND literal

{\small
\begin{dlvcode}
\bigwedge S = \bigwedge\{ p(X): a(X,Y),b(X)\}
\end{dlvcode}
}

Then the instantiation of $S$ is

{\small
\begin{dlvcode}
S' = \{\tuple{p(1):a(1,1),b(1)},\tuple{p(1):a(1,2),b(1)},\\\quad \quad \ \ \tuple{p(2):a(2,1),b(2)}, \tuple{p(2):a(2,2),b(2)}\}
\end{dlvcode}
}

and its value w.r.t $I$ is $I(S')$ = \{$ p(1), p(2)$\}.
$\bigwedge S'$ is true w.r.t. $I$ because both $p(1)$ and $p(2)$ are true w.r.t $I$.
\end{example}

Let $r$ be a ground rule in $ground( \p )$.  The head of $r$ is {\em
true} w.r.t. $I$ if at least one literal of $H(r)$ is true w.r.t $I$. The body of $r$ is
{\em true} w.r.t. $I$ if all body literals of $r$ are true w.r.t. $I$.
The rule $r$ is {\em satisfied} (or
{\em true}) w.r.t.  $I$ if its head is true w.r.t. $I$ or its body is
false w.r.t. $I$.

A {\em model} for $\p$ is an interpretation $M$ for $\p$ such that
every rule $r \in ground(\p)$ is true w.r.t. $M$.  A model $M$ for
$\p$ is {\em minimal} if no model $N$ for $\p$ exists such that $N$
is a proper subset of $M$.
%The set of all minimal models for $\p$ is
% denoted by ${\rm MM}(\p )$.

{\bf Answer Sets.}
First we define the answer sets of standard positive programs (i.e. without parametric literals).
Then, we give a reduction from
full \dlpp programs (i.e. containing negation as failure and parametric literals) to standard
positive programs. Such a reduction is used to define answer
sets of \dlpp programs.

An interpretation $I \subseteq \BP$ is called {\em closed
under \p} (where \p{} is a positive standard program
without parametric literals), if, for every $\R \in \GP$, $\HR
\cap I \neq \emptyset$ whenever $\BR \subseteq I$.
An interpretation $I \subseteq \BP$ is an {\em answer set} for a
standard positive program \p{}, if it is
minimal (under set inclusion) among all interpretations that are
closed under \p{}.%
\footnote{\label{fn}Note that we only consider {\em consistent
answer sets}, while in \cite{gelf-lifs-91} also the inconsistent set
of all possible literals can be a valid answer set.}

\begin{example}
\em
The positive program
{%\small
\begin{dlvcode}
  a \Or{} b \Or{} c.
\end{dlvcode}
}
has the answer sets $\{{a}\}$, $\{{b}\}$, and
$\{{c}\}$. The program
{%\small
\begin{dlvcode}
 a \Or{} b \Or{} c. \\
\vspace{3pt}
  \derives{} a.
\end{dlvcode}
}
has the answer sets
$\{b\}$ and $\{c\}$.
Finally, the positive program
{%\small
\begin{dlvcode}
  a \Or{} b \Or{} c.
  \\
  \vspace{3pt}  \derives{} a.\\
  \vspace{3pt} b \derives{} c.\\
 \vspace{3pt} c \derives{} b.
\end{dlvcode}
}
has the single answer set
the set $\{{b},{c}\}$.
\end{example}

We next extend the notion of {\em Gelfond-Lifschitz transformation}\cite{gelf-lifs-91} to \dlpp\ programs.
To this end, we introduce a new transformation $\delta$.

Given a set $F=\{f_1,\cdots,f_n\}$ of ground literals, we define the following transformation
$\delta$:
{\small
\begin{dlvcode}
\delta(\small \bigvee F) = f_1 \Or \cdots, \Or f_n \ \ \ \delta(\bigwedge F) = f_1, \cdots, f_n \end{dlvcode}
}
The {\em reduct} or {\em Gelfond-Lifschitz transform} of a
\dlpp program \p{} w.r.t.\ a set $I \subseteq \BP$ is the positive
ground program $\p^I$, obtained from $Ground(\p{})$ by
the following steps:
\begin{enumerate}
\item
Replace each instance $\bigvee S'$ of a parametric OR literal $\bigvee S$ by $\delta(\bigvee I(S'))$.
\item
Replace each instance $\bigwedge S'$ of a parametric AND literal $\bigwedge S$ by $\delta(\bigwedge I(S'))$.

\item Delete all rules $\R \in \p$ for which a negative literal in
      $\BR$ is false w.r.t. $I$.
\item Delete the negative literals
      from the remaining rules.
\end{enumerate}

An answer set of a program \p{} is a set $I \subseteq \BP$ such
that $I$ is an answer set of $\GP^I$.

\begin{example}
\em

Consider the following
\dlpp program $\p_1$
{\small
\begin{dlvcode}
p(1). \ \ \ a(1).  \ \ \ a(2). \vspace{3pt}\\
q \derives \bigwedge\{not\ p(X):a(X)\}.
\end{dlvcode}
}
and $I = \{p(1) \Comma a(1) \Comma a(2)\}$.
The instantiation of the set $\{not\ p(X):a(X)\}$
is {\small\begin{dlvcode}S' = \{\tuple{not\ p(1): a(1)}, \tuple{not\ p(2): a(2)}\}.\end{dlvcode}}
By evaluating $S'$ w.r.t $I$ we obtain {\small\begin{dlvcode}I(S') = \{not\ p(1),  \ not\ p(2) \}\end{dlvcode}}
Now, by applying step (2) of the the reduct we obtain the program
{\small
\begin{dlvcode}
p(1). \ \ \ a(1).  \ \ \ a(2). \vspace{3pt}\\
q \derives not\ p(1), not\ p(2).
\end{dlvcode}
}
and then, by applying step (3) we delete the rule, as $not \ p(1)$ is false, obtaining
{\small
\begin{dlvcode}
\p_1^I=  \{ p(1). \Comma a(1). \Comma a(2). \}.
\end{dlvcode}
}
Obviously, $I$ is an answer set of $\p_1^I$ and then,
it is also an answer set for $\p_1$.
\\

Now, consider the program $\p_2$
{\small
\begin{dlvcode}
  p(1). \vspace{3pt}\\

  \bigvee \{b(X):a(X)\}.\\
  \end{dlvcode}
  }
and $J = \{p(1)\}$. We have that the instantiation of $\{b(X):a(X)\}$ is
{\small\begin{dlvcode}S' = \{\tuple{b(1): a(1)}\}.\end{dlvcode}}
and $J(S') = \emptyset$.
By applying step (1) of the reduct, we obtain an empty disjunction which evaluates false
in any interpretation.
Then, the reduct $\p_2^J$ has no answer sets and so $J$ it is not an answer set of $\p_2$.
Note that $\p_2$ has no answer sets.
\end{example}

\section{Declarative Programming in Standard DLP}\label{sec:gcc}
\subsection{The \gc Declarative Programming Methodology}
\label{methodology}

The standard \DLP language can be used to encode problems in a
highly declarative fashion, following a ``\gc'' (Guess/Check)
paradigm. In this section, we will describe this technique and we
then illustrate how to apply it on a number of examples.  Many
problems, also problems of comparatively high computational
complexity (that is, even $\SigmaP{2}$-complete problems), can be
solved in a natural manner with \DLP by using this declarative
programming technique. The power of disjunctive rules allows for
expressing problems which are even more complex than \NP, and the
(optional) separation of a fixed, non-ground program from an input
database allows to do so uniformly over varying instances.

Given a set $\factprog{}_I$ of facts that specify an instance $I$ of some
problem ${\bf P}$, a \gc program $\aprog{}$ for ${\bf P}$ consists of the following
two main parts:

\begin{description}
\item[Guessing Part] The guessing part $\guessprog{} \subseteq \aprog{}$ of the
  program defines the search space, in a way such that answer sets of
  $\guessprog{} \cup \factprog{}_I$ represent ``solution candidates'' for $I$.

\item[Checking Part] The checking part $\checkprog{} \subseteq \aprog{}$ of the
  program tests whether a solution candidate is in fact an admissible solution,
  such that the answer sets of
  $\guessprog{}\cup \checkprog{} \cup \factprog{}_I$ represent the
  solutions for the problem instance $I$.

\end{description}

The two layers above can also use additional auxiliary predicates,
which can be seen as a background knowledge.

In general, we may allow both $\guessprog{}$  and
$\checkprog{}$ to be arbitrary collections
of rules in the
program,  and it may  depend on  the complexity  of the  problem which
kinds of rules  are needed to realize these  parts (in particular, the
checking  part); we defer  this discussion  to a  later point  in this
chapter.

Without  imposing  restrictions  on  which  rules  $\guessprog{}$  and
$\checkprog{}$  may  contain,  in  the  extremal  case  we  might  set
$\guessprog{}$ to  the full program  and let $\checkprog{}$  be empty,
i.e.,  all checking  is integrated  into the  guessing part  such that
solution  candidates are  always  solutions. However,  in general  the
generation of the search space may  be guarded by some rules, and such
rules might  be considered more  appropriately placed in  the guessing
part  than in  the checking  part.  We do  not pursue  this issue  any
further here, and thus also refrain from giving a formal definition of
how to separate a program into a guessing and a checking part.

For many problems, however, a natural \gc program can be designed, in
which the two parts are clearly identifiable and have a simple
structure:

\begin{itemize}
\item The guessing part $\guessprog{}$ consists of some disjunctive rules which
  ``guess'' a solution candidate~$S$.
\item The checking part $\checkprog{}$ consists of integrity constraints which
  check the admissibility of $S$.
\end{itemize}
All two layers may also use additional auxiliary predicates, which are
defined by normal stratified rules.
Such auxiliary
predicates may also be associated with the guess for a candidate, and
defined in terms of other guessed predicates, leading to a more ``educated
guess'' which reduces blind guessing of auxiliary predicates; this
will be seen in some examples below.

Thus, the disjunctive rules define the search space in which rule
applications are branching points, while the integrity constraints
prune illegal branches.

{\bf Remark.} The \gc programming methodology has positive implications also from
the Software Engineering viewpoint.
Indeed, the modular program structure in \gc
allows us to develop programs incrementally providing support for
simpler testing and debugging activities.
Indeed, one first writes the Guess module $\guessprog{}$
and tests that $\guessprog{}\cup \factprog{}_I$ correctly
defines the search space.
Then, one deals with the Check module and verifies that the answer sets of
$\guessprog{} \cup \checkprog{}\cup \factprog{}_I$ are the admissible
problem solutions.

%%%%%%%%%%%%%%%%%%%%%%%%%%%%%%%%%%%%%%%%%%%%%%%%%%%%%%%%%%%%
\subsection{Applications of the \gc Programming Technique}

In this section, we illustrate the declarative programming
methodology described in Section~\ref{methodology}
by showing its application on a couple of standard problems from graph theory.
%%%%%%%%%%%%%%%%%%%%%%%%%%%%%%%%%%%%%%%%

\subsubsection{Hamiltonian Path}\label{sec:HAMPATH}

Consider now a classical
\NP-complete problem in graph theory, namely {\em Hamiltonian Path}.

\begin{definition}[HAMPATH]
Given a directed graph $G=(V,E)$ and a node $a \in V$ of this graph,
does there exist a path of $G$ starting at $a$ and passing through
each node in $V$ exactly once?
\end{definition}

Suppose that the graph $G$ is specified by using predicates
\code{node} (unary) and \code{arc} (binary), and the starting node
is specified by the predicate \code{start} (unary).
Then, the following \gc program \aprog[hp] solves the problem HAMPATH.
{\small
\[
\begin{array}{l@{}l}
\begin{array}{l}
\hspace{-0.7cm}\code{\dlrule{inPath(X,Y) \Or{} outPath(X,Y)\\}{start(X), arc(X,Y)}}\>\\
\hspace{-0.7cm}\code{\dlrule{inPath(X,Y) \Or{} outPath(X,Y)\\}{reached(X), arc(X,Y)}}\>\\
\end{array}
&
\left.
\begin{array}{@{}l@{}}
~\\
~\\
~
\end{array}
\hspace{-0.3cm}\right\}\ \mbox{\textbf{Guess}}\\[2ex]
\begin{array}{ll}
\hspace{-0.7cm} \code{\dlconstraint{inPath(X,Y), inPath(X,Y1), Y <> Y1}}
 \\
\hspace{-0.7cm} \code{\dlconstraint{inPath(X,Y), inPath(X1,Y), X <> X1}}
 \\
\hspace{-0.7cm} \code{\dlconstraint{node(X), \naf\ reached(X), \naf\ start(X)}}
\end{array}
&
\left.
\begin{array}{@{}l@{}}
~\\
~\\
~
\end{array}
\hspace{-0.3cm}\right\} \ \mbox{\textbf{Check}}\\[2ex]
\begin{array}{l}
%\code{\dlrule{reached(X)}{start(X)}}\\
\hspace{-0.7cm}\code{\dlrule{reached(X)}{inPath(Y,X)}}
\end{array}
&
\left.
\begin{array}{@{}l@{}}
~\\
~
\end{array}
\hspace{-0.3cm}\right\}\
\begin{array}{l}
\mbox{\textbf{Auxiliary}}\\
\mbox{\textbf{Predicate}}
\end{array}
\end{array}
\]
}
The two disjunctive rules guess a subset $S$ of the given arcs to be
in the path, while the rest of the program checks whether that subset
$S$ constitutes a Hamiltonian Path. Here, an auxiliary predicate
$\code{reached}$ is used, which is associated with the guessed
predicate $\code{inPath}$ using the last rule.
%.. is used by both
%the guessing part $\guessprog{}$ and the checking part $\checkprog{}$.

The predicate $\code{reached}$ influences through the second
rule the guess of $\code{inPath}$, which is made somehow inductively:
Initially, a guess on an arc leaving the starting node is made by the
first rule, and then a guess on an arc leaving from a reached node by
the second rule, which is repeated until all reached nodes are
treated.

In the Checking Part, the first two constraints
check whether the set of arcs $S$ selected by \code{inPath} meets
the following requirements, which any Hamiltonian Path must satisfy:
(i) there must not be two arcs starting at the same node, and
(ii) there must not be two arcs ending in the same node.
The third constraint enforces that all nodes in the graph
are reached from the starting node in the subgraph induced by
$S$. This constraint also ensures that this subgraph is connected.

It is easy to see that any set of arcs $S$ which satisfies all
three constraints must contain the arcs of a path
$v_0,v_1,\ldots,v_k$ in $G$ that starts at node $v_0=a$, and passes
through distinct nodes until no further node is left, or it arrives at
the starting node $a$ again. In the latter case, this means that the
path is a Hamiltonian Cycle, and by dropping the last arc, we
have a Hamiltonian Path.

Thus, given a set of facts $\factprog{}$ for $node$, $arc$, and $start$,
specifying the problem input, the program $\aprog[hp]\cup \factprog{}$ has an answer
set if and only if the input graph has a Hamiltonian Path.
Thus, the above program correctly encodes the decision problem
of deciding whether a given graph admits an Hamiltonian Path or not.

This encoding is very flexible, and can be easily adapted to solve
both the {\em search problems} Hamiltonian Path and Hamiltonian Cycle
(where the result is to be a tour, i.e., a closed path).  If we want
to be sure that the computed result is an {\em open} path (i.e., it is
not a cycle), then we can easily impose openness by adding a further
constraint \code{\dlconstraint{start(Y), inPath(\_,Y)}} to the program
(like in Prolog, the symbol `$\code{\_}$' stands for an
anonymous variable, whose value is of no interest).
Then, the set $S$ of selected arcs in an answer set of $\aprog[hp] \cup \factprog{}$
constitutes a Hamiltonian Path starting at $a$.
If, on the other hand, we want to compute a Hamiltonian Cycle,
then we have just to strip off the literal $\naf\ start(X)$
from the last constraint of the program.

%%%%%%%%%%%%%%%%%%%%%%%%%%%%%%%%%%%%%%%%
\subsubsection{N-Coloring}\label{sec:NCOL}
Now we consider another classical \NP-complete problem from graph
theory, namely {\em N-Coloring}.

\begin{definition}[N-COLORING]
Given a graph $G=(V,E)$, a N-Coloring of $G$ is an assignment of
one, among N colors, to each vertex in $V$, in such a way that
every pair of vertices joined by an edge in $E$ have different
colors.
\end{definition}
Suppose that the graph $G$ is
represented by a set of facts with predicates \code{vertex} (unary) and \code{edge} (binary),
respectively. Then, the following \DLP
program $\gp_{col}$ determines the admissible ways of coloring
the given graph.
{\small
\[
\begin{array}{l@{}l}
\begin{array}{l}
\hspace{-0.5cm}\code{\dlrule{col(X,I) \Or{} not\_col(X,I)}{\\vertex(X), color(I)}}\\
\end{array}
& \left.
\begin{array}{@{}l@{}}
~\\
~
\end{array}
\right\}\ \mbox{\textbf{Guess}}\\[2ex]
\begin{array}{ll}
\hspace{-0.2cm} \code{\dlconstraint{col(X,I), col(Y,I), edge(X,Y)}}
 \\
\hspace{-0.2cm} \code{\dlconstraint{col(X,I), col(X,J), I <> J}}
\\
\hspace{-0.2cm} \code{\dlconstraint{vertex(X), \naf \ colored(X)}}

\end{array}
& \left.
\begin{array}{@{}l@{}}
~\\
~\\
~
\end{array}
\right\} \ \mbox{\textbf{Check}}\\[2ex]
\begin{array}{l}
%\code{\dlrule{reached(X)}{start(X)}}\\
\hspace{-0.5cm}\code{\dlrule{colored(X)}{col(X,I)}}
\end{array}
& \left.
\begin{array}{@{}l@{}}
~\\
~
\end{array}
\right\}\
\begin{array}{l}
\mbox{\textbf{Auxiliary}}\\
\mbox{\textbf{Predicate}}
\end{array}
\end{array}
\]
}
$col(X,I)$ says
that vertex $X$ is assigned to color $I$ and $not\_col(X,I)$ that
it is not. The disjunctive rule guesses a graph coloring; the constraints in the checking part verify that the
guessed coloring is a legal N-Coloring. In particular the first constraint asserts that
two joined vertices cannot have the same
color, while the remaining two constraints impose that each vertex is
assigned to exactly one color.

The answer sets of $\gp_{col}$  are all the possible
legal N-Colorings of the graph. That is, there is a one-to-one
correspondence between the solutions of the N-Coloring problem and
the answer sets of $\gp_{col}$. The graph
is N-colorable if and only if there exists one of such
answer sets.

\subsubsection{Maximal Independent Set}\label{sec:INDSET}
Another classical problem in graph theory is the independent set problem.

\begin{definition}[Maximal Independent Set]
Let $G=(V,E)$ be an undirected graph, and let $I\subseteq V$. The
set $I$ is {\em independent} if whenever $i,j \ \in I$ then there
are no edges between $i$ and $j$ in $E$. An independent set $I$ is {\em
maximal} if no superset of $I$ is an independent set.
\end{definition}

Suppose that the graph $G$ is represented by a set of facts $F$ with predicates
\code{node} (unary) and \code{edge} (binary). The following program $\gp_{IndSet}$
computes the maximal independent sets of $G$:
{\small
\[
\begin{array}{l@{}l}
\begin{array}{l}
\hspace{-0.7cm}\code{(r_1) \ \ \dlrule{in(X) \Or{} out(X)}{node(X)}}\\
\end{array}
& \left.
\begin{array}{@{}l@{}}
~\\
~
\end{array}
\right\}\ \mbox{\textbf{Guess}}\\[2ex]
\begin{array}{ll}
\hspace{-0.7cm} \code{(c_1) \ \ \dlconstraint{in(X), in(Y), edge(X,Y)}}\\
\hspace{-0.7cm} \code{(c_2) \ \ \dlconstraint{out(X),  not \ toBeExcluded(X)}}\\
 \end{array}
& \left.
\begin{array}{@{}l@{}}
~\\
~\\
~
\end{array}
\right\} \ \mbox{\textbf{Check}}\\[2ex]
\begin{array}{l}
%\code{\dlrule{reached(X)}{start(X)}}\\
\hspace{-0.7cm}\code{(r_2) \ \ \dlrule{toBeExcluded(X)}{in(Y), edge(X,Y)}}
\end{array}
& \left.
\begin{array}{@{}l@{}}
~\\
~
\end{array}
\right\}\
\begin{array}{l}
\mbox{\textbf{Auxiliary}}\\
\mbox{\textbf{Predicate}}
\end{array}
\end{array}
\]
}
The rule $r_1$ guesses a set of vertices;
$in(X)$ means that node $X$ belongs to the set while $out(X)$
means that it does not.
Then, the integrity constraint $c_1$ verifies that the guessed set is independent.
In particular, it says that it is not possible that two nodes joined by an edge belong to the set.

Note that the answer sets of $F\cup\{r_1, c_1\}$ correspond exactly to the independent sets of $G$.

The maximality of the set is enforced by constraint $c_2$ using the auxiliary predicate $toBeExcluded$.
A node $X$ has to be excluded by the set because a node connected to it is already in the set.
Then $c_2$ says that it is not possible that a node is out of the set if there is no reason to exclude it.

\section{Knowledge Representation by \dlpp}\label{sec:kr}
In this section, we show how \DLP extended by parametric connectives can be used to
encode relevant problems in a natural and elegant way.

\nop{
\newcounter{ind}
\def\nbc{\pr{ind}}
\[
\begin{array}{l@{\hspace*{0.5cm}}rcl}
\nbc & in(X) & \leftarrow & node(X) , not\ notin(X)\\
\nbc & notin(X) & \leftarrow & edge(X,Y), in(Y)\\
\end{array}
\]
}
\subsection{N-Coloring}\label{sec:N-Coloring}

In the previous section we showed an encoding for the N-Coloring problem,
following the \gc paradigm.
Now, we show how the extension of \DLP with parametric connectives allows us to represent the N-Coloring problem
in a much more intuitive way by simply modifying the elegant encoding of 3-colorability described in the Introduction.

Suppose again that the graph in input is represented by
 predicates
\code{vertex} (unary) and \code{edge} (binary) and the set of $N$ admissible colors is provided by a set of facts
$color(c_1), \cdots, color(c_N)$.
Then, the following \dlpp program computes the N-Colorings of the graph.
{\small
\[
\begin{array}{l}
\code{\dlrule{ (r) \ \ \bigvee\{col(X,C):  color(C) \}}{vertex(X)}}\\
\code{(c) \ \ \dlconstraint{  \ \ col(X,C), col(Y,C), edge(X,Y), X \neq Y}}
\end{array}
\]
}
Rule (r) guesses all possible N-Colorings. It contains in
the head a parametric literal representing the disjunction of all the
atoms $col(X,c_1), \cdots,$ $col(X,c_N)$, where $c_1, \cdots, c_N$
are the $N$ colors (i.e. the disjunction of all the atoms
representing the possible ways to color $X$). For each vertex $v$,
the following ground rule belongs to the instantiation of the
program:
{\small
\begin{dlvcode}
\hspace{-0.7cm}\bigvee\{\tuple{col(v,c_1):color(c_1)}, \cdots, \tuple{col(v,c_N):color(c_N)}\}\\
\derives vertex(v).
\end{dlvcode}
}
Since vertex $v$ and $color(c_1), \cdots, color(c_N)$ are always true, the above rule stands for the following disjunction
{\small
\[
\code{col(v,c_1) \vee \cdots \vee col(v,c_N)}
\]
}
The integrity constraint (c) simply checks that the N-Coloring is correct, that
is, adjacent nodes must always have different colors.

\subsection{Maximal Independent Set}
Another problem which can be easily encoded in a more intuitive
way by \dlpp is maximal independent set shown in section \ref{sec:INDSET}. Indeed, this problem can
be represented by the following {\em one-rule} encoding.
{\small
\[
\begin{array}{l}
\code{\dlrule{in(X)}{node(X), \bigwedge\{\naf \ in(Y):
arc(X,Y)\}}}\\
\end{array}
\]
}
As usual, the graph in input is encoded by predicates $node$ and $arc$ and the atom $in(X)$ means that
node $X$ belongs to the set.
Intuitively, such rule says that node $X$ belongs to the independent set if,
for each node $Y$ which is connected to it, $Y$ does not belong to the set.
In particular, the parametric AND literal \code{\bigwedge\{not \ in(Y):
arc(X,Y)\}} is the conjunction of all the literals $not \ in(Y)$ such that
there exists an edge between $X$ and $Y$.

Note that, differently from the \gc encoding shown in the previous section
this formulation does not need the predicate $out(X)$ and the auxiliary predicate $toBeExcluded(X)$
used to mark the nodes that have to be excluded by the set.

It is worth noting that we do not need further rules to express
maximality property, which, indeed, comes for free.

\subsection{N-Queens}\label{sec:NQUEENS}
We next illustrate a \dlpp encoding of the well-known N-Queens problem.

\begin{definition}[N-QUEENS]
Place N queens on a N*N chess board such that the placement of no queen constitutes an attack on any other.
A queen attacks another if it is in the same row, in the same column, or on a diagonal.

\end{definition}

Suppose that rows and columns are represented by means of
facts $row(1). , \cdots, row(N).$ and $column(1). , \cdots,
column(N).$ Then the following \dlpp program solves the N-Queens
problem.
\[
\small
\begin{array}{l}
\code{\dlrule{\bigvee\{q(X,Y):  column(Y) \}}{row(X)}}\\
\code{(c_1) \ \ \ \dlconstraint{q(X,Y), q(Z,Y), X \neq Z}}\\
\code{ (c_2) \ \ \ \dlconstraint{q(X1,Y1), q(X2,Y2), \\
 \hspace{1.2 cm}X2=X1+K, Y2=Y1+K, K > 0}}\\
\code{(c_3) \ \  \ \dlconstraint{q(X1,Y1), q(X2,Y2), \\
\hspace{1.2 cm} X2=X1+K, Y1=Y2+K, K > 0}}\\
\end{array}
\]

We represent queens by atoms of the form $q(X,Y)$. $q(X,Y)$ is true if a queen
is placed in the chess board at row $X$ and column $Y$.
The disjunctive rule guesses the position of the queens; in particular,
for each row $X$, we guess the column where the queen has to be placed.
Then the constraints assert that two queens cannot stay in the same column (constraint $c_1$)
and in the same diagonal (from top left to bottom right (constraint $c_2$) and from top right to bottom left (constraint $c_3$)).

\section{Implementation Issues}\label{sec:implementation}
In this section we illustrate the design of the implementation of the parametric connectives in
the \dlv\ system.
We first recall the architecture of \dlv\, and we then discuss the impact of the implementation of parametric connectives in \dlv.

\subsection{\dlv\ Architecture}
\begin{figure}
\begin{center}
\epsfig{file=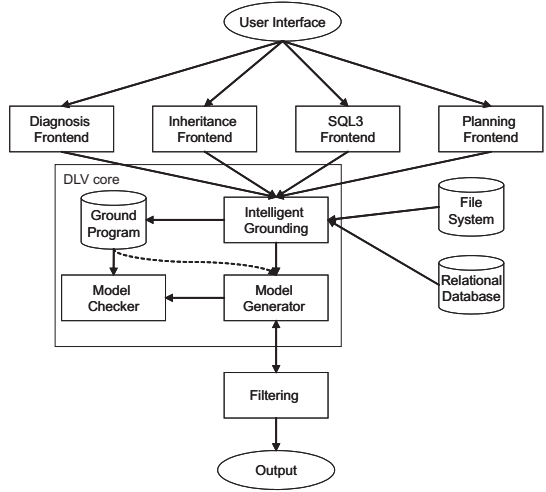,height=6.5cm}
\end{center}
\HELP
\caption{The System Architecture of \dlv}
\label{fig:architecture}
\HELP
\end{figure}

An outline of the general architecture of the \dlv\ system is depicted in
\refFig{architecture}.
The general flow in this picture is top-down. The principal User
Interface is command-line oriented, but also a Graphical User
Interface (GUI) for the core systems and most front-ends is
available. Subsequently, front-end transformations might be
performed. Input data can be supplied by regular
files, and also by relational databases. The \dlv\ core then produces answer sets one at a time, and
each time an
answer set is found, ``Filtering'' is invoked, which performs post-processing
(dependent on the active front-ends) and controls continuation or
abortion of the computation.

The \dlv\ core consists of three major components: the ``Intelligent
Grounding,'' the ``Model Generator,'' and the ``Model Checker'' modules that share
a principal data structure, the ``Ground Program''. It is created by
the Intelligent Grounding using differential (and other advanced) database techniques together
with suitable data structures, and used by the Model Generator and
the Model Checker. The Ground Program is guaranteed to have exactly the same answer sets as the original program. For some syntactically restricted classes of programs (e.g.\ stratified programs), the Intelligent Grounding module already computes
the corresponding answer sets.

For harder problems, most of the computation is performed by the Model
Generator and the Model Checker.  Roughly, the former
produces some ``candidate'' answer sets (models) \cite{fabe-etal-99b,fabe-etal-2001a}, the stability
and minimality of which are subsequently verified by the latter.

The Model Checker (MC) verifies whether the model at hand is an
answer set.  This task is very hard in general, because checking the
stability of a model is known to be co-\NP-complete. However, MC
exploits the fact that minimal model checking --- the hardest part --- can be efficiently performed
for the relevant class of \emph{head-cycle-free} (HCF) programs.

\subsection{Efficient Implementation of Parametric Connectives in \dlv}
Implementing the full \dlpp language in the \dlv\ system, would have a strong impact on \dlv\ requiring
many changes to all modules of the \dlv\ core,
including the Model Generator (MG) and the Model Checker (MC).
Making such changes would increase the complexity of the code and it could lead to an efficiency loss,
because, besides the standard literals, a new kind of literals, should be manipulated.
In order to obtain an efficient implementation, we impose a syntactic restriction
on the domain predicates (i.e. on the predicates appearing in the conjunction
on the right side of symbolic sets) that allows us to translate parametric literals into standard conjunctions and disjunctions
during the instantiation. In this way, the grounding produces standard \DLP programs and no changes to Model Generator
and Model Checker are necessary.

In particular, we impose that such predicates are normal
(disjunction-free) and stratified \cite{apt-etal-88}. For each symbolic set $S=\{L:Conj\}$, all domain literals of $S$ in $Conj$
are instantiated \underline{before} than dealing with the parameter $L$.
Thus, when the symbolic set $S$ has to be grounded all the domain predicates of $S$ are fully instantiated
and ready to be used.
Thanks to the imposed restrictions on the domain predicates (which are normal, stratified predicates), their truth values
are fully decided, that is they are either true or false.
Consequently, we can limit the instantiation of $S$ only to the ``useful'' atoms, that is, the instances of $L$ such that
the corresponding instances of $Conj$ are true.
\begin{example}\em
Consider the program

{\small
\begin{dlvcode}
a(1). \ \ a(2).  \ \ a(3). \ \ a(4). \ \ c(1).
\vspace{3pt}\\
b(X) \derives a(X), \  not \ c(X).
\vspace{3pt}\\
\bigvee\{p(X):b(X)\}.
\end{dlvcode}
}
The grounding procedure first instantiates the rule $b(X) \derives a(X), \ not \ c(X)$,
and generates the instances $b(2), \ b(3), \ b(4)$ for the domain predicate $b$.
Next, it considers $\bigvee\{p(X):b(X)\}.$ generating the standard disjunction
$p(2) \Or p(3) \Or p(4)$.
\end{example}
\section{Related Work}\label{sec:relwork}
We are not aware of other proposals for extending \DLP by parametric connectives. However,
our work has some similarity with other extensions of logic programming by other forms of nested operators like
for instance the nested expressions defined in \cite{lifs-etal-99}.

Our parametric disjunction has some similarity also with weight constraints of Smodels \cite{simo-etal-2002}.

A weight constraint is an expression of the form
$l\{L:D\}u$. The integer numbers $l$ and $u$ represent the lower and the upper bound of the constraint,  respectively.
$L:D$ is called conditional literal, $L$ is a standard literal and the conditional part $D$ is a domain predicate which is required
to be normal and stratified.
Thus, the parametric OR literal {\small \begin{dlvcode}\bigvee\{col(X,C):  col(C) \}\end{dlvcode}}  is similar to the
Smodels weight constraint
{\small \begin{dlvcode}1\{col(X,C):  col(C) \}1\end{dlvcode}}
However, it is worthwhile noting that the above Smodels construct derives exactly one atom
while the semantics of \dlpp follows the standard interpretation of disjunction (at least one atom is derived).
For instance, the \dlpp program
{\small
\begin{dlvcode}
c(1). \ \  \ c(2). \vspace{3pt}\\
\bigvee\{a(X): c(X)\}. \vspace{3pt}\\
a(1) \derives a(2).\vspace{3pt}\\
a(2) \derives a(1).
\end{dlvcode}
}
has the single answer set $\{ a(1)., \ a(2)., \ c(1)., \ c(2).\}$
Contrariously, the Smodels program
{\small
\begin{dlvcode}
c(1). \ \  \ c(2). \vspace{3pt}\\
1\{a(X): c(X)\}1 \vspace{3pt}\\
a(1) \derives a(2).\vspace{3pt}\\
a(2) \derives a(1).
\end{dlvcode}
}
has no answer sets.
\section{Conclusions}\label{sec:conclusion}

We have proposed \dlpp, an extension of \DLP by parametric connectives.
These connectives allow us to represent compactly the disjunction/conjunction
of a set of atoms having a given property enhancing the knowledge modelling abilities of \DLP.

We have formally defined the semantics of the new language, and
we have shown the usefulness of \dlpp on relevant
knowledge-based problems.

Ongoing work concerns the implementation of parametric literals in the \dlv\ system
following the design presented in section \ref{sec:implementation}.
Moreover, we are analyzing also the computational complexity of \dlpp which interestingly seems
to be the same as for standard \DLP.
Further work concerns an experimentation activity devoted to the evaluation of the impact of parametric
connectives on system efficiency. We believe that the conciseness of the encoding obtained
through parametric literals in some cases, like for instance N-Coloring and N-Queens, should bring
a positive gain on the efficiency of the evaluation.

\section*{Acknowledgments}

%\vspace{-.2cm}
This work was supported by the European Commission under projects INFOMIX,
 IST-2002-33570 INFOMIX, IST-2001-32429 ICONS, and IST-2001-37004 WASP.

%We described the (ongoing) implementation
%of these constructs in the \dlv\system.
%%%%%%%%%%%%%%%%%%%%%%%%%%%%%%%%%%%%%%%%%%%%%%%%%%%%%%%%%%%%%%%%%%%%%%%%%%%%%%%
\bibliography{../bibtex/bibtex}
\bibliographystyle{abbrv}

\end{document}